\documentclass[11pt,onecolumn,twoside]{IEEEtran}

\usepackage[utf8]{inputenc} % allow utf-8 input
\usepackage{amssymb}       % blackboard math symbols
\usepackage{graphicx}
\usepackage{amsthm}
\usepackage{amsmath}
\usepackage{dsfont}

\newtheorem{theorem}{Theorem}
\newtheorem*{theorem*}{Theorem}
\newtheorem{definition}{Definition}
\newtheorem*{definition*}{Definition}
\newtheorem{corollary}{Corollary}[theorem]
\newtheorem*{corollary*}{Corollary}
\newtheorem{lemma}{Lemma}
\newtheorem*{lemma*}{Lemma}

\newtheorem*{prop*}{Proposition}

\DeclareMathOperator*{\argmin}{argmin}

\begin{document}
\title{Online Reinforcement Learning with Passive Memory}

\author{Anay Pattanaik and Lav R.\ Varshney% <-this % stops a space
\thanks{A.\ Pattanaik is with the Department of Computer Science and Coordinated Science Laboratory, University of Illinois Urbana-Champaign, USA (e-mail: anayp2@illinois.edu).}%
\thanks{L.~R.\ Varshney is with the Department of Electrical and Computer Engineering and Coordinated Science Laboratory, University of Illinois Urbana-Champaign, USA (e-mail: varshney@illinois.edu).}
\thanks{This work is supported in part by the National Science Foundation through Expeditions Grant IIS-2123781.
}%
}

\maketitle
\begin{abstract}
This paper considers an online reinforcement learning algorithm that leverages pre-collected data (passive memory) from the environment for online interaction. We show that using passive memory improves performance and further provide theoretical guarantees for regret that turns out to be near-minimax optimal. Results show that the quality of passive memory determines sub-optimality of the incurred regret. The proposed approach and results hold in both continuous and discrete state-action spaces.  
\end{abstract}

\section{Introduction}
Several reinforcement learning (RL) algorithms such as Q-learning \cite{sutton2018reinforcement}, deterministic policy gradient (DPG) \cite{silver2014deterministic}, trust region policy optimization (TRPO) \cite{schulman2015trust}, and proximal policy optimization (PPO) \cite{schulman2017proximal} do not leverage any memory or past interaction with the environment. However, the success of several deep learning RL counterparts such as deep Q-learning (DQN) \cite{mnih2015human}, deep double-Q learning (DDQN) \cite{van2016deep}, and deep deterministic policy gradient (DDPG) \cite{lillicrap2015continuous}) can be attributed in part to the \emph{replay buffer} that stores past interactions with the environment. The replay buffer is used to update the \emph{target network} of the value function and was first proposed in \cite{mnih2015human}. Based on empirical evidence, the authors conjecture that the replay buffer and target network stabilize training. In this paper, we take a step towards analyzing the theoretical properties of memory and its impact on online RL algorithms. We show that the incurred regret depends on the quality of passive memory.

In neuroscience and psychology, \emph{episodic memory} is defined as the ability to recall and mentally re-experience specific episodes from one’s personal past and is contrasted with \emph{semantic memory} that includes memory for generic, context-free knowledge \cite{GILLUND201268}. Moreover, \emph{associative memory} refers to forming relationships between potentially unrelated objects \cite{suzuki2005associative}, e.g.\ association of an aroma to a particular person or object. It has been shown that humans and other animals use episodic memory for decision-making \cite{GershmanRL} . The global neuronal workspace (GNW) theory of consciousness \cite{zacks2023evolutionary} and unlimited associative learning (UAL) \cite{birch2020unlimited} both argue that sensory perception, motor control, value, and event memory have evolutionary origin. 

The class of offline RL algorithms typically do not interact with the environment  \cite{levine2020offline}. This is in contrast to ``ideal'' human behavior where episodic memory is used for decision making. These algorithms also rely on density ratio realizability assumptions that may not hold in general for a given function class \cite{chen2022offline}. In contrast, our approach requires density coverage with respect to only the optimal policy.  Recent work on RL with prior data \cite{ball2023efficient} is complimentary to our work, as it provides several heuristics (design choices) with ablation studies to constrain the online interaction to offline data such as symmetric sampling of online and offline collected data. Unfortunately, \cite{ball2023efficient} does not provide any theoretical guarantees. Our proposed algorithm is more principled and provides theoretical guarantees that it is near-minimax optimal.

In the context of RL, \cite{Zhu2020Episodic} proposed an algorithm that leverages associative memory. However, this algorithm breaks down for simple stochastic environments and does not have any finite-sample theoretical guarantees; the convergence guarantee for Q values is asymptotic (rather than finite-sample) and is also limited to discrete sample spaces. Since people, animals, and numerous other systems operate in the continuous domain, it is preferable to have RL algorithms that can handle both discrete and continuous state-action spaces. Our work provides this flexibility and also comes with finite-sample guarantees.

Our approach builds upon the regularized linear programming (LP) formulation of RL and incorporates (passive) memory to improve the performance of RL, cf.~\cite{nachum2019algaedice}. Note that prior work \cite{nachum2019algaedice} uses a policy optimization approach based on unbiased policy gradients to find the optimal solution, showing impressive empirical performance. That work, however, does not provide any convergence or optimality guarantee, and observes that the Lagrangian form leads to numerical instability in practice. In contrast, we use an online learning approach (rather than a policy gradient approach) and provide the associated regret and performance guarantees. Our contributions for online RL with passive memory can be succinctly stated as follows.
\begin{itemize}
    \item We provide performance analysis for (regularized) LP formulation of RL with passive memory. Performance depends upon the \emph{quality} of passive memory, as given in Thm.~\ref{thm:perf_diff}.
    \item We provide the minimax regret lower bound for (discounted) online RL with finite horizon per episode, as in Thm.~\ref{thm:low_bd}.
    \item We provide a regret upper bound for online RL with passive memory for arbitrary density approximator, as in Thm.~\ref{thm:reg_upp_bd}.
    \item We provide a regret upper bound with plug-in and kernel density approximation for online RL with passive memory which leads to near-minimax optimalilty, as in Thm.~\ref{thm:reg_upp_cont} and Cor.~\ref{cor:reg_upp_disc}. Our approach also generalizes seamlessly to continuous state-action spaces.
\end{itemize}

The rest of paper is organized as follows: Section \ref{sec:background} provides the necessary background for online RL, Section \ref{sec:analysis} provides the theoretical analysis where we show that the proposed algorithm is near-minimax optimal, and Section \ref{sec:conc} gives concluding remarks and possible future directions. Most proofs are deferred to the Appendix.

\section{Background, Notation, and Algorithm}
\label{sec:background}

We briefly review the necessary background and establish notation for Markov decision process (MDP).  Then we introduce the regularized LP formulation of RL with passive memory, the algorithm we consider.
\subsubsection{(Discounted) Markov Decision Process} 
	\begin{itemize}
		\item MDP setting: A (discounted) Markov decision process is given by a tuple $(\mathcal{S},\mathcal{A},R,T,\gamma)$ where
		$\mathcal{S}$ is the state space, $\mathcal{A}$ is the action space, $R:\mathcal{S}\times\mathcal{A}\rightarrow \mathbb{R}$ is the reward function, $\gamma \in (0,1)$ is the discount factor, and $T$ is the transition kernel, where $T:\mathcal{S}\times\mathcal{A} \rightarrow \Delta(\mathcal{S})$ maps to a distribution over the state space $\Delta(\mathcal{S})$.
		\item Policy: A policy is a stochastic mapping from the state space to a distribution over the action space, that is, $\pi:\mathcal{S} \rightarrow \Delta(A)$. The optimal policy is denoted by $\pi^*$.
		\item Value function: The value function of state $s_0$ for policy $\pi$ is the discounted sum of future rewards if we start from $s_0$ and follow $\pi$, that is, $V^{\pi}(s_0)=\mathbb{E}[\sum_{t=0}^{\infty}\gamma^tR(s_t,a_t)|s_0,\pi]$. The optimal value function is usually denoted by $V^*(s_0)$.
		\item Value function for an initial state distribution: Similar to the value function for a state, we can define the value function for an initial state distribution as $V^{\pi}(\mu_0)=\mathbb{E}[\mathbb{E}[\sum_{t=0}^{\infty}\gamma^tR(s_t,a_t)|s_0,\pi]]$,  where $\mu_0$ is the initial state distribution.
		\item Discounted occupancy measure: We define discounted occupancy measure as the discounted measure over the state-action space if we start with an initial measure of $\mu_0$ and follow the policy $\pi$, that is,   $d^{\pi}_{\mu_0}(s,a)=(1-\gamma)\sum_{t=0}^{\infty}\gamma^t\mathbb{P}(s_t=s,a_t=a|\pi,\mu_0)$. We overload the notation and define $d(s,a):=d^{\pi}_{\mu_0}(s,a)$.
		\item Passive Memory: The distribution induced by passively collected data is given as $d^{\mathcal{D}}(s,a)$. 
		\item Regret: The regret for any policy $\pi$ is the difference in performance between the optimal policy and $\pi$, that is, $R_T$:= $\sum_{t=0}^TV^*(\mu_0)-V^{\pi_t}(\mu_0)$. 
	\item Goal: Throughout this work, the goal is to minimize regret.	
	\item Divergences: The Kullback-Leibler (KL) divergence is denoted by $D( d_1 \Vert d_2 )$ and the Bregman divergence with a generating function $\phi$ is denoted by $D_\phi (x,y)$. 
	\end{itemize}
To make the manuscript less cumbersome, we have taken the liberty to denote densities and measures with the same notation, but this should be clear from context.
 
\subsubsection{Regularized LP formulation of RL}
The LP formulation can be given as \cite{puterman2014markov}:  
\begin{align*}
V^{*}(\mu_0)&=\max_{d}\mathbb{E}[R(s,a)]\\
 s.t. ~~ \int_{\mathcal{A}} d(s,a)&=(1-\gamma)\mu_0(s)+\gamma T_{\mbox{*}}d(s), \forall s\\
\text{where} ~~T_{\mbox{*}}(ds)&:=\int_{\mathcal{S} \times \mathcal{A}}T(ds|s',a')d(s',a').
\end{align*}

The regularized dual with KL divergence is:
\begin{align}\label{eq:lp_kl}
V^{\tilde{\pi}}(\mu_0)&=\max_{d}\mathbb{E}[R(s,a)]-D(d||d^{\mathcal{D}})\nonumber\\
 s.t. ~~ \int_{\mathcal{A}} d(s,a)&=(1-\gamma)\mu_0(s)+\gamma T_{\mbox{*}}d(s), \forall s.
\end{align}

The conjugate of the dual and its optimal solution is given as \cite{nachum2020reinforcement}:
\begin{align}
\tilde{V}^*&=\argmin_V(1-\gamma)\mathbb{E}_{\mu_0}[V(s_0)] +\log \mathbb{E}_{d^{\mathcal{D}}}[\exp\{R(s,a)+\gamma T V(s,a)-V(s)\}],\label{eq:lp_dual_opt}
\end{align}
\begin{align*}
\tilde{d}(s,a)=\frac{d^{\mathcal{D}}(s,a)\exp(R+\gamma T \tilde{V}^*-\tilde{V}^* (s,a))}{\int_{\mathcal{S} \times \mathcal{A}}d^{\mathcal{D}}(s',a')\exp(R+\gamma T \tilde{V}^*-\tilde{V}^* (s',a'))}.
\end{align*}
Here $T V(s,a):=\mathbb{E}_{s' \sim T(s,a)}[V(s')]$.

The corresponding optimal policy can be obtained by 
\begin{align*}
\tilde{\pi}(a|s)=\frac{\tilde{d}(s,a)}{\int_{\mathcal{A}}\tilde{d}(s,a')}.
\end{align*}

\section{Online RL and Analysis}\label{sec:analysis}
In this section, we analyze the performance of online RL with passive memory and also provide the regret lower and upper bounds for online RL. We show that the proposed approach is near-optimal. Throughout the paper, we will assume without loss of generality that the reward is normalized to lie in $[0,1]$. We also assume that the state and action space is either compact or discrete, and the measure of entire state space is absolutely continuous (w.r.t.\ some underlying $\sigma$-finite measure on the Borel $\sigma$-algebra) bounded by $S$, $A$, respectively. 

\subsection{Performance with Passive Memory}
We first characterize the suboptimality of the proposed algorithm for a given dataset $d^{\mathcal{D}}(s,a)$ for an MDP $\mathcal{M}$. 
\begin{theorem}[Performance difference analysis]\label{thm:perf_diff}
	Given a dataset $d^{\mathcal{D}}(s,a)$ for an MDP $\mathcal{M}$, the difference in the performance of the optimal policy and the policy produced by regularized LP formulation of RL is given by
\begin{align*}
V^{\pi^*}(\mu_0)-V^{\pi}(\mu_0) \leq \sqrt{\log\left\Vert\frac{d^*}{d^\mathcal{D}}\right\Vert_{\infty}(1-\gamma)cSA}.
\end{align*}
Here $d^*(s,a)$ is the state-action distribution induced by the optimal policy, $c=\frac{1}{\left\Vert\frac{\mu_0}{d^{\mathcal{D}}}\right\Vert_{-\infty}}$, and $||\cdot||_{-\infty}$ is the short hand notation for $\min_{s}|f(s)|$. Further, $S, A$ are the bounds on a compact state and action space, respectively.	
\end{theorem}
\begin{IEEEproof} The proof is provided in Appendix  \ref{thm:perf_diff_app}.
\end{IEEEproof}

\subsection{Regret Lower Bound}
Next we consider the regret lower bound. We will build two similar MDPs where good performance in one MDP leads to bad performance in the other MDP. Let the first MDP be denoted by $\mathcal{M}=\langle\mathcal{S},\mathcal{A},\mathcal{R}, T \rangle$ and the other MDP be denoted by $\mathcal{M}'=\langle\mathcal{S},\mathcal{A},\mathcal{R}', T \rangle$. Let a policy $\pi$ be run in $\mathcal{M}'$ and $\mathcal{M}$ and the history be denoted by $H=(s_0,a_0,r_0,s_1,a_1,r_1,\ldots,s_H,a_H,r_H)$. The following lemma will be useful.
\begin{lemma}\label{lem:kl_hist}
	Let $P_{\pi,M}$ and $P_{\pi,M'}$ be the (discounted) distribution of reward induced by policy $\pi$ in the MDPs $\mathcal{M}$ and $\mathcal{M}'$, respectively. Then the divergence between $P_{\pi,M}$ and $P_{\pi,M'}$ can be expressed as
	\begin{align*}
	D(P_{\pi,M}||P_{\pi,M'})=\int_{\mathcal{S}\times\mathcal{A}}\mathbb{E}[T_{s,a}(H)]D(d_{s,a}||d'_{s,a})\,ds\,da.
	\end{align*}
	Here $T_{s,a}(H)$ is the discounted time spent in state-action $(s,a)$ when the episode is truncated after a horizon of length $H$, and $d_{s,a}$ represents the distribution of reward at state $(s,a)$ for MDP $\mathcal{M}$. 
\end{lemma}
\begin{IEEEproof}
The proof is provided in Appendix \ref{lem:kl_hist_app}.    
\end{IEEEproof} 

Lemma \ref{lem:kl_hist} lets us relate the divergence between the long-term expected reward with the expected time spent in each state-action pair and the divergence between the distribution of rewards for the different MDPs $\mathcal{M}$ and $\mathcal{M}'$.

If there are $T$ policies and $n$ episodes are repeated for each policy for a finite horizon of length $H$, then the minimax lower bound for the regret can be given as follows.
\begin{theorem}[Minimax regret lower bound]\label{thm:low_bd}
	The minimax lower bound for continuous state-action space where the state space is a bounded compact space with bounded measure $S$ and the action space is also a bounded compact space with bounded measure $A$ is: 
	\begin{align*}
	R_T = \Omega\left(\sqrt{\frac{1+\gamma^H}{1-\gamma}SAnT}\right).
	\end{align*}
	Here $H$ is the length of truncated horizon, $n$ is the number of iterations used for density estimation for each episode by an agent, and $T$ is the number of episodes.
\end{theorem}
\begin{IEEEproof}
The proof is provided in Appendix \ref{thm:low_bd_app}.
\end{IEEEproof}
Observe that the discrete case can easily be derived as a corollary by using discrete measure in Theorem~\ref{thm:low_bd}.

\subsection{Regret Upper Bound}
Next we consider the regret upper bound for online RL that uses passive memory. We will first establish a general regret upper bound given an (imperfect) density estimation oracle. We will then use kernel density estimation techniques for the induced distribution over continuous state-action space, and plug-in density estimator for discrete state-action space in place of the oracle.
\begin{theorem}[Regret upper bound with density estimation error]\label{thm:reg_upp_bd}
	Given passive memory $d^{\mathcal{D}}$ and a density estimation oracle with  $\epsilon$ as the density estimation error, then the regret can be upper bounded by
	\begin{align*}		
	R_T=\mathcal{O}\left(\sqrt{D(d^*,d^{\mathcal{D}})SA\left(\epsilon+\frac{\gamma^{H}}{1-\gamma}\right)nT}\right).
	\end{align*}
	Furthermore, if the passively collected data has support over the occupancy measure of the optimal policy,
	\begin{align}
	R_T=\mathcal{O}\left(\sqrt{nTSA\log (SA)\left(\epsilon+\frac{\gamma^{H}}{1-\gamma}\right)}\right).\label{eq:reg_upp_bd_apr}
	\end{align}
	Here $S$, $A$ are the measures of compact state-action spaces and $H$ is the truncated horizon length.
\end{theorem}
\begin{IEEEproof}
The proof is provided in Appendix \ref{thm:reg_upp_bd_app}.    
\end{IEEEproof}
Note that for a discrete domain, the measures can be replaced by discrete measures. Thus in Theorem~\ref{thm:reg_upp_bd}, the measures of compact space can directly be replaced by the cardinality of $\mathcal{S}$ and $\mathcal{A}$. 

We can see from Theorem~\ref{thm:reg_upp_bd} that if the passively collected data is ``close'' to the optimal distribution, then we incur a low regret with online RL algorithm. 

We consider two separate cases for density estimation error---the plug-in estimator for discrete action spaces and kernel density estimator for continuous domains---and use those estimation errors to find a probably approximately correct (PAC) regret bound.
\begin{lemma}[Plug-in density estimator]\label{lem:den_plug}
	If we use the following plug-in density estimator for discrete state action space 
		 $$\hat{d}^H(s,a)=\frac{(1-\gamma)}{n(1-\gamma^H)}\sum_{i=1}^n\left(\sum_{h=0}^H\gamma^h\mathbf{1}[s_h^i,a_h^i=s,a] \right),$$ then with probability at least $1-\delta$
		\begin{align*}
		||\hat{d}^H-d||_{\infty}\leq \frac{\frac{\log \bar{\delta}}{3}+\sqrt{\frac{\log^2\bar{\delta}}{9}+8\log\bar{\delta}}}{2n},
		\end{align*}
		where $\bar{\delta}=\frac{2|\mathcal{S}||\mathcal{A}|}{\delta}$.
\end{lemma}
\begin{IEEEproof}
The proof is provided in Appendix \ref{lem:den_plug_app}.
\end{IEEEproof}
Lemma \ref{lem:den_plug} enables us to upper-bound the regret when we use a plug-in density estimator.

Next, we focus on the state-action density for continuous domains and use the following kernel density estimator.
 \begin{definition}[State-action kernel density estimator]
 	Let the kernel density estimate be 
 	\begin{align*}
 	\hat{d}(s,a)=\frac{(1-\gamma)}{n(1-\gamma^H)}\sum_{i=1}^n\frac{1}{b^{d}}\sum_{h=0}^H\gamma^hK\left(\frac{(s,a)-(s_h^n,a_h^n)}{b}\right).
 	\end{align*}
 \end{definition} 
We will make standard assumptions on the properties of the kernel function \cite{tsybakov2009introduction}, namely:
\begin{enumerate}
	\item $d(s,a)\in\Sigma(\beta,L)$, here $\Sigma(\beta,L)$ is the H\"{o}lder class, that is, $\Sigma(\beta,L):=\{g: |D^sg(x)-D^sg(y)|\leq L||x-y||$ for all $s$ such that $|s|=\beta-1\}$ and $D^s$ is the $s$th partial derivative. For example, $d=1,\beta=1$ gives the Lipschitz property for the derivative, that is, $|g'(x)-g'(y)|\leq L||x-y||$. 
	\item If $g\in\Sigma(\beta,L)$, then $|g(x)-g_{x,\beta}(u)|\leq L|x-y|$ where $g_{x,\beta}(u)=\sum_{|s|\leq\beta}\frac{(u-x)^s}{s!}D^sg(x)$.
	\item $K(x)=G(x_1)G(x_2)\cdots G(x_d)$ where $G$ has support on $[-1,1]$, $\int G =1$, $\int |G|^p<\infty$ for all $p\geq1$ and $\int t^sK(t)\,dt=0$ for all $s\leq\beta$.
	\item $\int |t|^\beta|K(t)|\,dt<C_{K}$ and $\int t^sK(t)\,dt=0$, for all $s\leq\beta$.
\end{enumerate}
An example of a kernel that satisfies the above condition for $\beta=2$ is $K(x)=0.75(1-x^2), |x|\leq1$. Next we calculate the bias of the proposed estimator.
\begin{lemma}\label{lem:sup_est_error}
	Let $\bar{d}(s,a):=\mathbb{E}[\hat{d}(s,a)]$, then the bias of the estimator of the occupancy density in a H\"{o}lder class can be bounded as
	\begin{align*}
	\sup_{d \in \Sigma(\beta,L)}|d(s,a)-\bar{d}(s,a)|\leq LC_Kb^\beta.
	\end{align*}	
\end{lemma}
\begin{IEEEproof}
The proof in provided in Appendix \ref{lem:sup_est_error_app}.    
\end{IEEEproof}

Once we know the bias of our estimator, we decompose the upper bound on the $L_1$ error appropriately using McDiarmid's inequality \cite{boucheron2013concentration}. 
\begin{lemma}\label{lem:l1_est_error}
	If $\bar{d}(s,a):=\mathbb{E}[\hat{d}(s,a)]$, then we have that
	with probability at least $1-\delta$,
	\begin{align*}
	||\hat{d}-\bar{d}||_{L1} \leq LC_Kb^{\beta}SA+\sqrt{\frac{\ln \frac{1}{\delta}}{2nb^{2d}}}.
	\end{align*} 
\end{lemma}
\begin{IEEEproof}
The proof is provided in Appendix \ref{lem:l1_est_error_app}.
\end{IEEEproof}

Combining Lemma \ref{lem:l1_est_error} with the generalized regret bound of Thm.~\ref{thm:reg_upp_bd} will complete the desired results.

\begin{theorem}[Regret upper bound for continuous case]\label{thm:reg_upp_cont}
	With probability at least $1-\delta$, we have that 
	\begin{align*}
	R_T =\mathcal{O}\left(\sqrt{nT\log(SA)SA\left(\sqrt{\frac{\ln\frac{1}{\delta}}{2nb^{2d}}}+\frac{\gamma^H}{1-\gamma}\right)}\right).
	\end{align*}
Here $S$, $A$ are the measures of compact state-action spaces, $b$ is the bandwidth parameter of the density kernel, $d$ is the combined dimensions of the state action space, $H$ is the truncated horizon, and $n$ is the number of episodes used for density estimation of occupancy measure for each policy.
\end{theorem}
\begin{IEEEproof}
We substitute the density approximation error in Lemma \ref{lem:l1_est_error} into the regret obtained in \eqref{eq:reg_upp_bd_apr} (with discrete measure) to obtain the required result if we run the same policy $n$ times for density estimation.
\end{IEEEproof}
\begin{corollary}[Regret upper bound for discrete case]\label{cor:reg_upp_disc}
With probability at least $1-\delta$, the regret 
\begin{align*}
	R_T = \mathcal{O}\left(\sqrt{nT\log |\mathcal{S}||\mathcal{A}|\left(\frac{|\mathcal{S}||\mathcal{A}|}{n}\log\frac{ |\mathcal{S}||\mathcal{A}|}{\delta}+\frac{\gamma^{H}}{1-\gamma}\right)}\right).
	\end{align*}
	Here $S$, $A$ are the discrete state-action spaces, $H$ is the truncated horizon, and $n$ is the number of episodes used for density estimation of the occupancy measure for each policy.
\end{corollary}
\begin{IEEEproof}
We substitute the density approximation error in Lemma \ref{lem:den_plug} into the regret obtained in  \eqref{eq:reg_upp_bd_apr} (with discrete measure) to obtain the required result if we run the same policy $n$ times for density estimation.
\end{IEEEproof}

\section{Conclusions and Future Work}\label{sec:conc}
We characterize the suboptimality of regularized LP formulation of RL. We also provide a passive memory based online RL algorithm and prove that it is nearly minimax optimal. Future work shall involve analyzing \emph{active} memory where we purge and fill the memory as needed during online interaction rather than relying passively on pre-collected data from the environment.

\bibliographystyle{IEEEtran}
\bibliography{references}

\appendix
\label{sec:appendix}

\subsection{Proof of Thm.~\ref{thm:perf_diff}}
\label{thm:perf_diff_app}

\begin{IEEEproof}
		We start by establishing the necessary condition for optimality for the LP of \eqref{eq:lp_dual_opt}.
		
\begin{align}
(1-\gamma)D_V\mathbb{E}_{\mu_0}\left[V\right]+D_V\log \mathbb{E}_{d^\mathcal{D}}\left[\exp\{R+\gamma TV-V \}\right]&=0\nonumber\\
(1-\gamma)\mu_0+\frac{d^{\mathcal{D}}(\gamma -1 )\mathbb{E}\left[\exp\{R+\gamma TV-V \}\vert S\right]}{\mathbb{E}\left[\exp\{R+\gamma TV-V \}\right]}&=0\nonumber\\
\frac{\mathbb{E}\left[\exp\{R+\gamma TV-V \}\vert S\right]}{\mathbb{E}\left[\exp\{R+\gamma TV-V \}\right]}=\frac{\mu_0}{d^{\mathcal{D}}}\nonumber\\
\frac{\lVert\exp\{R+\gamma TV-V \}\rVert_{\infty}}{\mathbb{E}\left[\exp\{R+\gamma TV-V \}\right]}\geq\left\Vert\frac{\mu_0}{d^{\mathcal{D}}}\right\Vert_{-\infty}\label{eq:bd_big_qt}
\end{align}		
Where we used an unusual notation $\left\Vert \cdot \right\Vert_{-\infty}$ to denote the minimum of absolute value, that is, $\min|\cdot|$.

We then use the necessary condition of \eqref{eq:bd_big_qt} to obtain the performance difference as follows. The difference between the optimal and regularized solution can be given by:
\begin{align*}
\left\vert\mathbb{E}_{d^*}R-\mathbb{E}_{\tilde{d}}R\right\vert
&\leq \left\Vert d^*-\tilde{d}\right\Vert_1 \left\Vert R\right\Vert_{\infty} \text{ (H\"older's inequality)}\\
&\leq \sqrt{\frac{1}{2}D\left(d^*\Vert\tilde{d}\right)} \text{ (Pinsker's inequality and $R$ is bounded)}\\
&=\sqrt{\frac{1}{2}D\left(d^*\Vert\frac{d^{\mathcal{D}}\exp\{R+\gamma TV-V \}}{\mathbb{E}\left[\exp\{R+\gamma TV-V \}\right]}\right)}\\
&\leq \sqrt{D\left(d^*\Vert d^{\mathcal{D}}\right)+\int_{\mathcal{S} \times \mathcal{A}}d^*\log\frac{\mathbb{E}\left[\exp\{R+\gamma TV-V \}\right]}{\exp\{R+\gamma TV-V \}}}\\
&\leq \sqrt{SA\log\left\Vert\frac{d^*}{d^\mathcal{D}}\right\Vert_{\infty}-SA\log \left\Vert\frac{\mu_0}{d^{\mathcal{D}}}\right\Vert_{-\infty}}
\text{(from necessary condition \eqref{eq:bd_big_qt})}\\
&\leq \sqrt{\log\left\Vert\frac{d^*}{d^\mathcal{D}}\right\Vert_{\infty}(1-\gamma)cSA}.
\end{align*}
\end{IEEEproof}

\subsection{Proof of Regret Bounds}
\subsubsection{Proof of Lemma \ref{lem:kl_hist}}
\label{lem:kl_hist_app}

\

\begin{IEEEproof}
	Note that using the Markov property, we can write the joint (discounted) distribution of a given history as $P_{\pi,M}(H)=(1-\gamma)\prod_{h=0}^{H}\gamma^h\pi(a_h|s_h)P(r|s_h,a_h)T(s_{h+1}|s_h,a_h)$. Further, the KL divergence can be expanded as
	\begin{align}
	D(P_{\pi,M}||P_{\pi,M'}) &=\sum_{h=1}^{H}\mathbb{E}\left[\mathbb{E}\left[D(P_{R_h}\Vert P'_{R_h})|S_h,A_h\right]\vert H^{h-1}\right]\nonumber\\
	&\text{(Chain rule; $R_h$ is reward at time $h$; $H^h$ is history till  $h$)}\nonumber\\
	&=\sum_{h=1}^{H}\mathbb{E}\left[\int_{\mathcal{S}\times\mathcal{A}}\delta_{(S_h,A_h)}(s,a)D(P_{R_h|s,a}\Vert P'_{R_h|s,a})\vert H^{h-1}\right]\nonumber\\
	&=\int_{\mathcal{S}\times\mathcal{A}}\sum_{h=1}^{H}\mathbb{E}\left[\delta_{(S_h,A_h)}(s,a)D(P_{R_h|s,a}\Vert P'_{R_h|s,a})\vert H^{h-1}\right]\nonumber\\
	&\text{(Fubini's Theorem)}\nonumber\\
	&=\int_{\mathcal{S}\times\mathcal{A}}D(P_{R_h|s,a}\Vert P'_{R_h|s,a})\mathbb{E}\left[\sum_{h=1}^{H}\delta_{(S_h,A_h)}(s,a)\vert H^{h-1}\right]\nonumber\\
	&\text{(Markovian reward)}\nonumber\\
	&=\int_{\mathcal{S}\times\mathcal{A}}D(d_{s,a}\Vert d'_{s,a})\mathbb{E}\left[\sum_{h=0}^H\delta_{(S_h,A_h)}(s,a)\right]\,d(s,a)\nonumber\\
	&=\int_{\mathcal{S}\times\mathcal{A}}D(d_{s,a}\Vert d'_{s,a})\mathbb{E}\left[T_{s,a}(H)\right]d(s,a).\nonumber
	\end{align}
\end{IEEEproof}

\subsubsection{Proof of Minimax Lower Bound Theorem \ref{thm:low_bd}}
\label{thm:low_bd_app}

\

\begin{IEEEproof}
	Let $\Delta \in (0,1)$, consider an MDP with reward function given by 
	\begin{align*}
	R(s,a) =
	\begin{cases}
	Ber(\frac{1}{2}+\Delta);  &(s_1,a_1)\\
	Ber(\frac{1}{2}); & \mbox{otherwise.}
	\end{cases}
	\end{align*}
	Let ${(\bar{s},\bar{a})=\arg min_{s,a}\mathbb{E}_{P_{\pi,M}}[T_{s,a}(H)]}$. Note that $${\int_{\mathcal{S}\times\mathcal{A}}\mathbb{E}_{P_{\pi,M}}[T_{(s,a)}(H)]\,ds\,da=\frac{nT}{1-\gamma}},$$ but $\int_{\mathcal{S}\times\mathcal{A}}\mathbb{E}_{P_{\pi,M}}[T_{(s,a)}(H)]\,ds\,da\geq \min_{s,a}\mathbb{E}_{H\sim P_{\pi,M}(.)}[T_{(s,a)}(H)]SA$. Thus, $\min_{s,a}\mathbb{E}_{P_{\pi,M}}[T_{(s,a)}(H)]\leq  \frac{nT}{(1-\gamma)SA}$.   Now consider another MDP $\mathcal{M}'$ with the same transition dynamics and state-action space but reward function given by 
	\begin{align*}
	R'(s,a)=
	\begin{cases}
	Ber(\frac{1}{2}+\Delta); &(s_1,a_1)\\
	Ber(\frac{1}{2}+2\Delta); &(\bar{s},\bar{a})\\
	Ber(\frac{1}{2});       & \mbox{otherwise.}
	\end{cases}
	\end{align*}
	Let $E$ be the event that $T_{s_1,a_1}\leq \frac{nT}{2(1-\gamma)}$. Now
	\begin{align*}
	R_T^{M}\mathds{1}_E&\geq\frac{nT\Delta}{2(1-\gamma)}\\
	R_T^{M'}\mathds{1}_{E^c}&\geq\frac{nT\Delta}{2(1-\gamma)}\\
	R_T^{M} + R_T^{M'}&\geq \frac{nT\Delta}{2(1-\gamma)}(\mathbb{P}_{\pi,M}(E)+\mathbb{P}_{\pi,M'}(E^c))\\
	&=\frac{nT\Delta}{2(1-\gamma)}(1+\mathbb{P}_{\pi,M}(E)-\mathbb{P}_{\pi,M'}(E))\\
	&\geq  \frac{nT\Delta}{2(1-\gamma)} (1-|\mathbb{P}_{\pi,M}(E)-\mathbb{P}_{\pi,M'}(E)|)\\
	&\qquad \text{(Used $-|a-b| \leq a-b$)}
	\end{align*}
	Now, 
	\begin{align*}
	|\mathbb{P}_{\pi,M}(E)-\mathbb{P}_{\pi,M'}(E)| &\leq D_{TV}({P}_{\pi,M},{P}_{\pi,M'})\\
	&\leq \sqrt{\frac{1}{2}D(P_{\pi,M}||P'_{\pi,M})}
	 \text{ (Pinsker's Inequality)}
	\end{align*}
	We also know that 
	\begin{align*}
	D(P_{\pi,M}||P'_{\pi,M})
	&\leq \sum_{s,a}\mathbb{E}[T_{s,a}(H)]D\left(Ber\left(\frac{1}{2}\right)||Ber\left(\frac{1}{2}+2\Delta\right)\right) 
	\text{ (from Lemma \ref{lem:kl_hist})}\\
	&\leq \frac{nT}{(1-\gamma)SA}O(\Delta^2)
	\text{ (using $D\left(Ber\left(\frac{1}{2}\right)||Ber\left(\frac{1}{2}+2\Delta\right)\right) = o(\Delta^2)$)} 
	\end{align*}
	Thus,
	\begin{align*}
	R_T^{M} + R_T^{M'} \geq \frac{nT\Delta}{2(1-\gamma)} \left(1-\sqrt{\frac{nT}{(1-\gamma)SA}c\Delta^2}\right).
	\end{align*}
	We can optimize w.r.t.\ $\Delta$ to get the optimal $\Delta=\sqrt{\frac{SA(1-\gamma)}{cnT}}$ and the regret $R_T^{M'}= \Omega\left(\sqrt{\frac{SAnT}{1-\gamma}}\right)$. Furthermore, the truncation of episode to $H$ steps gives an error of $\frac{\gamma^{H}}{1-\gamma}$ as shown in \eqref{eq:hor_trun}.
\end{IEEEproof}

\subsubsection{Proof of General Regret Upper Bound, Theorem \ref{thm:reg_upp_bd}}
\label{thm:reg_upp_bd_app}

\

\begin{IEEEproof}
	We observe the  regularized LP approach in \eqref{eq:lp_kl} through the lens of the online mirror descent (OMD) framework. Notice that the update equation is given as $\phi(\tilde{d}_{t+1})-\nabla \phi(d_t)=r$. Here $d_t$ is the discounted measure corresponding to the policy $\pi_t$. 
	\begin{align}
	V^*(\mu_0)-V^{\pi}(\mu_0)
	&=\langle d^*-d_t,r\rangle \nonumber\\
	&=\langle d^*-d_t, \nabla \phi(\tilde{d}_{t+1})-\nabla \phi(d_t)\rangle\nonumber\\
	&\leq D_{\phi}(d^*,d_t)+D_{\phi}(d_t,\tilde{d}_{t+1})-D_{\phi}(d^*,d_{t+1}) - D_{\phi}(d_{t+1},\tilde{d}_{t+1})\nonumber \text{ (Pythagoras theorem)}\nonumber\\
	R_T&\leq\sum_{t=0}^TD_{\phi}(d^*,d_t)-D_{\phi}(d^*,d_{t+1})+D_{\phi}(d_t,\tilde{d}_{t+1})\nonumber \\
	&\leq D_{\phi}(d^*,d^{\mathcal{D}})+\sum_{t=1}^TD_{\phi_{\mbox{*}}}(\nabla \phi(d_t)+r,\nabla\phi(d_t)) \text{ (using divergence equations of dual space)}\nonumber\\
	&= D_{\phi}(d^*,d^{\mathcal{D}})+\sum_{t=1}^T||r||_{w_t}^2 \nonumber\\
	& \text{(Taylor series expansion:}\nonumber\\
	&\quad\text{$\exists w_t$ with $\nabla\phi(w_t) \in [\nabla\phi(d_t), \nabla \phi(d_t)+r]$} \text{ and  $||r||^2_{w_t}:=r^T\nabla^2\phi_{\mbox{*}}(w_t)r$)}\nonumber\\
	&\leq D_{\phi}(d^*,d^{\mathcal{D}}) + \int_{\mathcal{S}\times\mathcal{A}}\sum_{t=1}^T\frac{r^2(s,a)\eta^2}{\phi''(d_t(s,a))}d(s,a) \text{ (using KL divergence)}\nonumber\\
	&\leq \frac{D(d^*,d^{\mathcal{D}})}{\eta} + \sum_{t=1}^T\int_{\mathcal{S}\times\mathcal{A}}\eta d_t(s,a)\,ds\,da
	\label{eq:regret_kl}
	\end{align}
	Then we incorporate horizon truncation into regret bound of \eqref{eq:regret_kl} as follows:
	\begin{align}
	\int_{\mathcal{S}\times\mathcal{A}}|d^H(s,a)-d(s,a)|\,ds\,da
	&=\int_{\mathcal{S}\times\mathcal{A}}\sum_{h=H+1}^\infty\gamma^hP(s_h,a_h=s,a)\,ds\,da\nonumber\\
	&= \frac{\gamma^{H}}{1-\gamma}\label{eq:hor_trun}
	\end{align}

 Using \eqref{eq:hor_trun} in \eqref{eq:regret_kl}, we get
	\begin{align*}
	R_T&\leq\sum_{t=1}^T\int_{\mathcal{S}\times\mathcal{A}} d_t^H(s,a)\eta+(d_t(s,a)-d_t^H(s,a))~\eta\,ds\,da  +\frac{D(d^*,d^{\mathcal{D}})}{\eta}\\
	&\text{($d_t^H$ is density approximation error for truncation)}\\
	&\leq \frac{D(d^*,d^{\mathcal{D}})}{\eta}+\sum_{t=1}^T\int_{\mathcal{S}\times\mathcal{A}} d_t^H(s,a)\eta\,ds\,da+\frac{T\eta\gamma^{H}}{1-\gamma} \text{ (from \eqref{eq:hor_trun})}\\
	&\leq\frac{D(d^*,d^{\mathcal{D}})}{\eta}+\sum_{t=1}^T\int \hat{d}_t^H(s,a)\eta\,ds\,da+(d_t^H(s,a)-\hat{d}_t^H(s,a))+\frac{T\eta\gamma^{H}}{1-\gamma}\\
	&\text{(assumed that we have an estimate of ${d}_{t-1}^H$ as $\hat{d}_{t-1}^H$)}\\
	&=\frac{D(d^*,d^{\mathcal{D}})}{\eta} +\eta\sum_{t=1}^T\int_{\mathcal{S}\times\mathcal{A}} \hat{d}_t^H(s,a)\eta\,ds\,da+\eta T\epsilon_1+\frac{T\eta\gamma^{H}}{1-\gamma}\\
	&\text{($\epsilon_1$ is the density estimation error)}\\
	&\leq \frac{D(d^*,d^{\mathcal{D}})}{\eta}+\eta TSA+\eta T\epsilon_1+\frac{T\eta\gamma^{H}}{1-\gamma}\\
	R_T&=\mathcal{O}\left(\frac{D(d^*,d^{\mathcal{D}})}{\eta}+\eta TSA\left(\epsilon+\frac{\gamma^{H}}{1-\gamma}\right)\right)\\
	\end{align*}
	By optimizing $\eta$, we obtain
	\begin{align*}
	R_T=\mathcal{O}\left(\sqrt{D(d^*,d^{\mathcal{D}})SA\left(\epsilon+\frac{\gamma^{H}}{1-\gamma}\right)T}\right)
	\end{align*} 
	Assuming that the passively collected data has support over the occupancy measure of the optimal policy, we can further upper bound $D_\phi(d^*,d^{\mathcal{D}})$ by  
	\begin{align}
	R_T=\mathcal{O}\left(\sqrt{SA\log SA\left(\epsilon+\frac{\gamma^{H}}{1-\gamma}\right)T}\right).
	\end{align}
\end{IEEEproof}

\subsubsection{Proof of plug-in density estimation Lemma \ref{lem:den_plug}}
\label{lem:den_plug_app}

\

\begin{IEEEproof}
	First, we observe that $\hat{d}^H(s,a)$ is an unbiased estimate of $d^H(s,a)$.
	\begin{align*}
	\hat{d}^H(s,a)&=\frac{(1-\gamma)}{n(1-\gamma^H)}\sum_{i=1}^n\left(\sum_{h=0}^H\gamma^h\mathbf{1}[s_h^i,a_h^i=s,a] \right)\\
	\mathbb{E}[\hat{d}^H(s,a)]&=\frac{(1-\gamma)}{n(1-\gamma^H)}\sum_{i=1}^n\left(\sum_{h=0}^H\gamma^h\mathbb{P}[s_h^i,a_h^i=s,a] \right)\\
	&= \frac{1}{n}\sum_{i=1}^nd^H(s,a)\\
	&=d^H(s,a)
	\end{align*}
	Let us find the variance of $\hat{d}^H(s,a)$:
	\begin{align*}
	&\mbox{Var}\left(\frac{(1-\gamma)}{(1-\gamma^H)}\sum_{h=0}^H\gamma^h\mathbf{1}[s_h,a_h=s,a] \right)=d^H(s,a)(1-d^H(s,a)).
	\end{align*}
	Now we invoke Bernstein's inequality \cite{bernstein1946theory} and get
	\begin{align*}
	\mathbb{P}(|\hat{d}^H-d^H|\geq \epsilon)&\leq 2\exp\left(-\frac{n\epsilon^2}{2(d^H(1-d^H)+\frac{\epsilon}{3})}\right)\\
	&\leq 2\exp\left(-\frac{n\epsilon^2}{2(1+\frac{\epsilon}{3})}\right)\text{(As $d^H(s,a)\leq 1$)}
	\end{align*}
	The union bound completes the proof as follows.
	\begin{align*}
	\mathbb{P}(||\hat{d}^H-d^H||_{\infty} >\epsilon)&=\mathbb{P}(\max_{s,a}|\hat{d}^H-d^H| >\epsilon)\\
	&\leq \sum_{s,a}\mathbb{P}(|\hat{d}^H-d^H| >\epsilon)\\
	&\leq \sum_{s,a}2\exp\left(-\frac{n\epsilon^2}{2(1+\frac{\epsilon}{3})}\right)\\
	&=|\mathcal{S}||\mathcal{A}|2\exp\left(-\frac{n\epsilon^2}{2(1+\frac{\epsilon}{3})}\right)\leq \delta
	\end{align*}
	so we obtain with probability at least $1-\delta$, that $$||\hat{d}^H-d^H||_{\infty}\leq \frac{\frac{\log \bar{\delta}}{3}+\sqrt{\frac{\log^2\bar{\delta}}{9}+8\log\bar{\delta}}}{2n}.$$
\end{IEEEproof}

\subsection{Proof of Lemma \ref{lem:sup_est_error}}
\label{lem:sup_est_error_app}

\

\begin{IEEEproof}

	\begin{align*}
	|\bar{d}(s,a)-d(s,a)|
	&=\left|\int C\sum_{h=0}^H\gamma^hK\left(\frac{(\bar{s},\bar{a})-(s,a)}{b}\right)d(\bar{s},\bar{a})\,d\bar{s}\,d\bar{a}-d(s,a)\right| \text{ (where $C=\frac{1-\gamma}{(1-\gamma^H)b^d}$)}\\
	&\leq \left|\int \frac{1}{b^d} K\left(\frac{\bar{x}-x}{b}\right)d(\bar{x})\,d\bar{x}-d(x)\right| \text{ (using $x:=(s,a)$ for convenience)}\\
	&=\left|\int K\left(||v||\right)d(x+bv)\,dv-d(x)\right| \text{(using $v:=\tfrac{\bar{x}-x}{b}$)}\\
	&=\left|\int K\left(||v||\right)d(x+bv)\,dv-\int K(||v||)d(x)\,dv\right| \text{ (as $\int K=1$)}\\
	&=\left|\int K\left(||v||\right)\{d(x+bv)-d(x)\}\,dv\right|\\
	&\leq \left|\int K\left(||v||\right)\{d(x+bv)-d_{x,\beta}(x+bv)\}\,dv\right|+\left|\int \{d(x)-d_{x,\beta}(x+bv)\}\,dv\right|\\
	&\leq Lb^\beta \int K(||v||)||v||^\beta +0 \text{ (using H\"{o}lder class property)}\\
	&\leq Lb^\beta C_K.
	\end{align*}
\end{IEEEproof}

\subsection{Proof of Lemma \ref{lem:l1_est_error}}
\label{lem:l1_est_error_app}

\

\begin{IEEEproof}
		We first show that the function $f(H^n)=||\hat{d}-\bar{d}||_{L_1}$ is a function of bounded difference, here $H^n$ represents a history length of $n$. So, if we perturb the $j$th observation to $j'$, the difference between the functions is
		\begin{align*}
		f(H^n)-f(H^n_{-j})
		&\leq C\int_{\mathcal{S}\times\mathcal{A}}\sum_{h=0}^H\gamma^h\left|K^j-K^{j'}\right|\,ds\,da\\
		&\text{where $C:=\frac{(1-\gamma)}{nb^{d}(1-\gamma^H)},K^j:=K\left(\frac{(s,a)-(s_h^j,a_h^j)}{b}\right)$}\\
		&\leq \frac{2}{nb^d}\int_{\mathcal{S}\times\mathcal{A}}K\left(\frac{(s,a)-(s_h^{j'},a_h^{j'})}{b}\right)\,ds\,da\\
		&=\frac{2}{nb^d}
		\end{align*}
		Now we apply McDiarmid's inequality \cite{boucheron2013concentration} to obtain
		\begin{align*}
		\mathbb{P}\left(\left| \Vert\bar{d}-\hat{d}\Vert_{L_1}-\mathbb{E}\left[\Vert\bar{d}-\hat{d}\Vert_{L_1}\right] \right|\right)&\leq 2\exp\left(\frac{-2\epsilon^2n^2b^{2d}}{4n}\right)\\
		&=2\exp\left(\frac{-\epsilon^2nb^{2d}}{2}\right).
		\end{align*}
		Thus, with probability at least $1-\delta$, we have $||\hat{d}-\bar{d}||_{L1} \leq\sqrt{\frac{\ln\frac{1}{\delta}}{2nb^{2d}}}+\mathbb{E}\left[\Vert\bar{d}-\hat{d}\Vert_{L_1}\right]$, but from Lemma \ref{lem:sup_est_error}, we get that  $\mathbb{E}\left[\Vert\bar{d}-\hat{d}\Vert_{L_1}\right] \leq LC_Kb^\beta SA$.
\end{IEEEproof}

\end{document}